\documentclass[sigconf]{acmart}
\usepackage{enumitem}
\AtBeginDocument{%
  }

\copyrightyear{2025}
\acmYear{2025}
\setcopyright{cc}
\setcctype{by}
\acmConference[UMAP Adjunct '25]{Adjunct Proceedings of the 33rd ACM Conference on User Modeling, Adaptation and Personalization}{June 16--19, 2025}{New York City, NY, USA}
\acmBooktitle{Adjunct Proceedings of the 33rd ACM Conference on User Modeling, Adaptation and Personalization (UMAP Adjunct '25), June 16--19, 2025, New York City, NY, USA}
\acmDOI{10.1145/3708319.3733659}
\acmISBN{979-8-4007-1399-6/2025/06}

\begin{document}


\title[LLMs and Social Choice Aggregation for Group Recommendations]{The Pitfalls of Growing Group Complexity: LLMs and Social Choice-Based Aggregation for Group Recommendations}


\author{Cedric Waterschoot}
\affiliation{%
  \institution{Maastricht University}
  \city{Maastricht}
  \country{The Netherlands}}
\email{cedric.waterschoot@maastrichtuniversity.nl}

\author{Nava Tintarev}
\affiliation{%
  \institution{Maastricht University}
  \city{Maastricht}
  \country{The Netherlands}}
\email{n.tintarev@maastrichtuniversity.nl}

\author{Francesco Barile}
\affiliation{%
  \institution{Maastricht University}
  \city{Maastricht}
  \country{The Netherlands}}
\email{f.barile@maastrichtuniversity.nl}

\renewcommand{\shortauthors}{Waterschoot et al.}

\begin{abstract}

Large Language Models (LLMs) are increasingly applied in recommender systems aimed at both individuals and groups. Previously, Group Recommender Systems (GRS) often used social choice-based aggregation strategies to derive a single recommendation based on the preferences of multiple people. In this paper, we investigate under which conditions language models can perform these strategies correctly based on zero-shot learning and analyse whether the formatting of the group scenario in the prompt affects accuracy. We specifically focused on the impact of group complexity (number of users and items), different LLMs, different prompting conditions, including In-Context learning or generating explanations, and the formatting of group preferences.
Our results show that performance starts to deteriorate when considering more than $100$ ratings. However, not all language models were equally sensitive to growing group complexity. Additionally, we showed that In-Context Learning (ICL) can significantly increase the performance at higher degrees of group complexity, while adding other prompt modifications, specifying domain cues or prompting for explanations, did not impact accuracy. We conclude that future research should include group complexity as a factor in GRS evaluation due to its effect on LLM performance. Furthermore, we showed that formatting the group scenarios differently, such as rating lists per user or per item, affected accuracy. All in all, our study implies that smaller LLMs are capable of generating group recommendations under the right conditions, making the case for using smaller models that require less computing power and costs.

\end{abstract}



\maketitle

\section{Introduction}

Group Recommender Systems (GRS) extend traditional recommender systems to obtain a single recommendation required to suit the preferences of a group of people. To process the potentially diverging ratings of multiple people, prior research has introduced social choice-based aggregation strategies \cite{masthoff2015group}. Recently, Large Language Models (LLMs) have been used to obtain group recommendations by providing them the individual ratings of group members \cite{Tommasel2024-dp,FENG2025agrouprec}. Social choice-based aggregation strategies were used as comparison for the LLM-generated recommendations. However, LLMs were not explicitly prompted to perform these strategies, even though strategies provide opportunities to adapt GRS to specific conditions and factors such as fairness and satisfaction \cite{barile2023evaluating,tran2019towards}. 
Understanding how the applied strategies align with, or differ from, LLM strategies helps us understand LLM capabilities for GRS better.

While recommender systems typically deal with large amounts of data, the number of total ratings the model has to consider, remains fairly overlooked. This complexity is important to consider since adding additional information to the prompt influences the output of LLMs \cite{Zamfirescu2023whyjohnny}. While the accuracy of traditional aggregation techniques for GRS are not affected by a growing number of items or group members, it is unclear how robust LLMs are against increasing group complexity. It is also unclear under which conditions this kind of robustness is present. Therefore, we formulate the following research questions:\newline

\noindent \textbf{RQ1.} Are LLMs robust against increasing group complexity, defined as the total number of ratings to consider, i.e. the product of group members and items, when applying aggregation strategies?

\noindent \textbf{RQ2.} Do additional prompt conditions; e.g., generating explanations, adding examples, or adding domain cues; impact the accuracy of LLMs when applying aggregation strategies?

\noindent \textbf{RQ3.} Does the format of the group; e.g., ratings per user or per item; impact the accuracy of LLMs when applying aggregation strategies?\newline

\begin{table*}[h]
  \caption{Social choice-based aggregation strategies derived from \cite{tran2019towards,barile2023evaluating,felfernig2018explanations,Waterschoot2025withfriends}}
  \label{tab:strats}
  \begin{tabular}{ccl}
    \toprule
    Strategy & Type& Procedure \\
    \midrule
    Additive Utilitarian (ADD) & Consensus & Recommends the item with the highest sum of all group members’ ratings \\
    \midrule
    Approval Voting (APP) & Majority &  Recommends the item with the highest number of ratings above a predefined threshold\\
    \midrule
    Least Misery (LMS) & Borderline & Recommends the item which has the highest of all lowest per-item
ratings\\
    Most Pleasure (MPL) & Borderline & Recommends the item with the
highest individual group member rating\\
    \bottomrule
  \end{tabular}
\end{table*}

To investigate these questions, we created $1,000$ fictitious groups with randomly $2, 4$, or $8$ members and $5, 10, 25$, or $50$ potential items to recommend. For each group, multiple LLMs were prompted to derive a group recommendation using a randomly chosen social choice-based aggregation strategy, presented to the LLM using a social choice-based explanation introduced in previous work \cite{najafian2018generating,kapcak2018tourexplain}. Furthermore, we explored several additional conditions that can affect LLM performance in generating correct recommendations at higher degrees of group complexity. We investigated whether prompting for natural language explanations affected the accuracy, a widespread use case of language models. Additionally, we tested whether adding domain cues (movie titles instead of anonymous items) or additional examples (in-context learning) improved the accuracy of recommendations generated by LLMs.
All in all, our paper makes the following contributions:

\begin{itemize}
    \item We highlight the importance of using a selection of LLMs due to uneven performance when generating group recommendations.
    \item We show that group complexity is an important factor in the evaluation of group recommendations generated by LLMs due to its effect on performance.
    \item We find that the prompting strategy of In-Context Learning significantly improves the capability of LLMs to apply social choice-based aggregation strategies at higher group complexity correctly.
    \item We show that the formatting of the group scenario impacts the accuracy of LLMs when applying aggregation strategies.
\end{itemize}

\section{Background}\label{sec:background}
In this section, we introduce the literature on Group Recommendation Systems (GRS) and social choice-based aggregation strategies. Additionally, we outline previous work on using LLMs for (group) recommendation and highlight the need for a more robust evaluation of LLM-generated recommendations in the context of standard aggregation methods for GRS.

\subsection{Group Recommendation}\label{sec:GRS}
While traditional recommender systems present output for a single user, GRS need to simultaneously process the preferences of multiple group members \cite{Masthoff2022group}. These systems are in rising demand and are being applied in various fields such as tourism \cite{chen2021attentive}, music \cite{najafian2018generating} and restaurant recommendation \cite{barile2021toward}. 

To derive a single recommendation which reflects the ratings of individual group members, individual ratings need to be aggregated. Inspired by \textit{Social Choice Theory} \cite{kelly2013social}, social choice-based aggregation strategies have been proposed to present a range of distinct options to process individual ratings into a single outcome \cite{masthoff2015group,masthoff2004group}. Even when more advanced approaches are proposed in which social factors or other dynamics influencing group decision-making are incorporated (see e.g. \cite{nguyen2019conflict,delic2018use}), these strategies are still widely used as procedure or baseline.

Broadly speaking, these strategies have been categorized as either \textit{consensus-based}, \textit{majority-based} or \textit{borderline} \cite{senot2010analysis}. In the current study, we make use of strategies representing each category, summarized in Table \ref{tab:strats}. The included consensus-based strategy, meaning that all ratings are considered, is \textit{Additive Utilitarian} (ADD), which recommends the item with the highest sum of all ratings in the group \cite{senot2010analysis}. As majority-based strategy, i.e., strategies that only make use of the most popular items or ratings, we included \textit{Approval voting} (APP). APP sets a threshold and recommends the item having the highest number of ratings above that threshold \cite{senot2010analysis}. Finally, two borderline strategies were included in the current study, as they consider diverging subsets of ratings. \textit{Least Misery} (LMS) recommends the item which has the highest of all lowest per-item rating, while \textit{Most Pleasure} (MPL) recommends the item with the highest overall rating by a group member \cite{senot2010analysis}.

\subsubsection{Social Choice-based Explanations}
To illustrate how social choice-based aggregation strategies work, previous studies introduced social choice-based explanations \cite{najafian2018generating,kapcak2018tourexplain}. These are natural language excerpts outlining the underlying mechanism of the strategy. A wide range of factors have been studied using this type of explanations, such as consensus perception regarding a group recommendation~\cite{barile2023evaluating,felfernig2018explanations}, privacy-preservation~\cite{najafian2021exploring,najafian2021factors} and fairness perception~\cite{tran2019towards,barile2023evaluating}. In the current study, we added a variation of social choice-based explanations to each prompt to instruct the LLM on the exact procedure necessary to obtain the correct group recommendation.

\subsection{LLMs for (Group) Recommendation}
Large Language Models (LLMs) have been increasingly implemented in the context of recommender systems \cite{Zhao2024rec}. Due to their interactive capabilities, LLMs have been used to address cold-start problems \cite{Sanner2023-kb,Wu2024could} or conversational recommender systems \cite{Gao2023chatrec,Yang2024behav}. In-Context Learning specifically, defined as the ability of LLMs to make predictions based on only a few training samples without updating parameters, has allowed LLMs to challenge conventional recommendation methodologies \cite{Hou2024large,dong2024survey}. Due to zero-shot and few-shot capabilities, LLMs have shown promise for data sparse tasks such as cross-domain recommendation \cite{Petruzzelli2024instruct,Kim2024large}. However, widespread use of LLMs, especially on external servers or making use of APIs, has raised privacy concerns, due to the prominence of user-related data in recommendation scenarios \cite{Wu2023enhanced,Zhao2024rec}.  An alternative presents itself in the use of smaller or open-source models that can run locally, omitting the need for external computation or API use \cite{Wiest2024privacy}.

The general performance of LLMs regarding group recommendation tasks remains understudied with recent work focusing on niche additions to the GRS pipeline, as opposed to a baseline accuracy of LLM-generated group recommendations. \citet{FENG2025agroup} used an LLM-based approach to mine topics from user comments of group members and use a Graph Convolution Network (GCN) to recommend items to the group based on those topics. Additionally, group recommendations generated by LLMs have been evaluated in terms of fairness and inclusion of sensitive attributes \cite{Tommasel2024-dp}. The authors evaluate movie recommendations made by three language models, with and without sensitive user attributes in the prompt, and compare it to a baseline by additive aggregation. However, although social choice-based aggregation strategies are commonly used to generate group recommendations, it is uncertain whether LLMs can effectively implement them. In this paper, we take a step back and investigate the accuracy of LLM-generated group recommendations and study the impact of prompt and data format on LLM-generated recommendations. These insights are a necessary component before the implementation of LLMs in the GRS pipeline and can guide future work on LLMs for GRS regarding prompt construction and group scenario formatting. Additionally, the influence of scenario complexity -- the total number of ratings to process -- has not been systematically examined in the evaluation pipeline. This is particularly important since group information is embedded within the prompt alongside other instructions and risk being forgotten, leading to an incorrect output. From this consideration, we decided to focus on the impact of group complexity, defined as the total number of ratings in the group scenario (based on the number of group members and the item set size).

\begin{figure*}[!t]
  \centering
  \includegraphics[width=.7\linewidth]{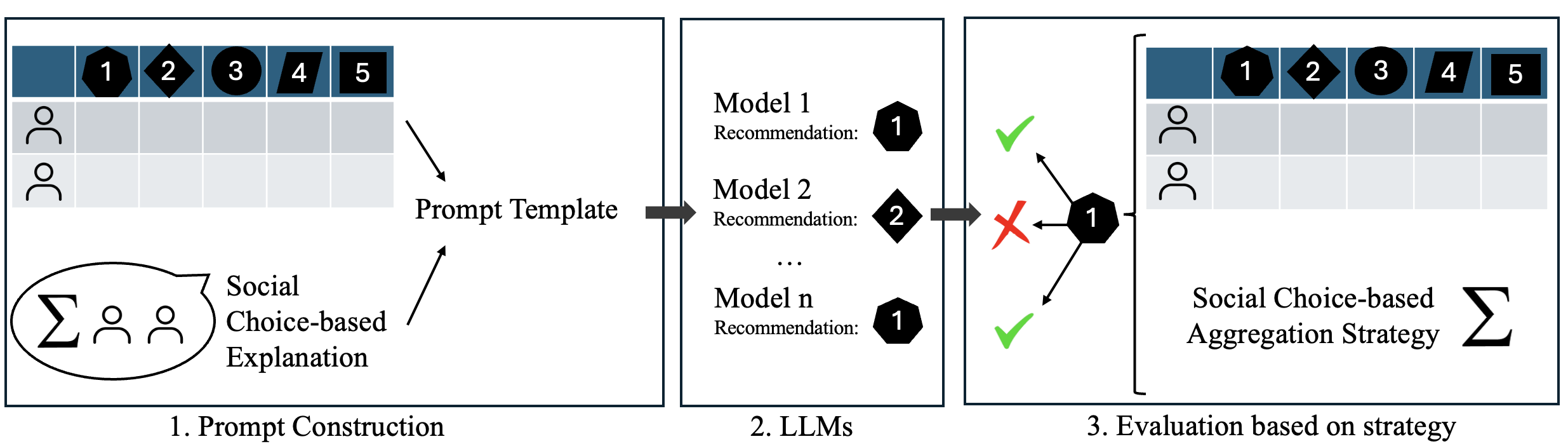}
  \caption{Pipeline for evaluating large language models' application of social choice-based aggregation strategies. We iterate through this pipeline for each given group scenario.}
  \label{fig:pipeline}
  \Description{Visualization of the pipeline. We construct a prompt to include group ratings and a social choice-based aggregation strategy}
\end{figure*}

\section{Methodology}\label{sec:methods}
In this section, we introduce the construction of fictitious group scenarios as well as the pipeline for applying social choice-based aggregation strategies using language models. After outlining the basic LLM procedure, we outline additional conditions that we applied to a selection of group scenarios.

\subsection{Group generation}\label{sec:groupgen}

In this study, we only focus on the aggregation step to check how proficient LLMs are in applying the strategies. Thus, we assume to have a rating for all options. These could be explicit ratings or estimated with individual recommender systems. For this analysis, we generated fictitious group scenarios. The procedure for generating a group matrix of size \textit{number of group members} ($group\_size$) x \textit{number of potential recommendations} ($num\_items$) was based on previous work \cite{barile2023evaluating,Waterschoot2025withfriends}. We adopted the code for generating random groups, not based on any (dis)similarity between members. As opposed to earlier work, we anonymized the items to omit any context or domain cues. Items were simply named $Item_x$, as opposed to ``$Rest_x$''. Additionally, users were represented with randomized IDs, not names. Per scenario, for each $Item_x$ with $x \in [1,num\_items]$ and each $User_y$ with $y \in [1,group\_size]$, a random rating between 0 and 10 was generated. The full code is found in the companion Github repository.\footnote{\url{https://github.com/Cwaterschoot/LLMs_SocialChoiceAggregation}}

We defined group complexity as the total number of ratings to be considered, i.e. the product between group size and the number of items. We randomly generated groups with either $2, 4$ or $8$ members ($group\_size$). The number of items presented per scenario were randomly generated at either $5, 10, 25$ or $50$ ($num\_items$). As a result, group complexity ranged from $10$, the most simplistic scenarios consisting of $2$ users and $5$ items, to $400$, representing complex groups with $8$ members and $50$ potential items to choose from. An example of a group scenario is presented in Table \ref{tab:groupscenario}. 

\begin{table}[h]
  \caption{Example of group scenario with group complexity = 10 (2 users x 5 items). Each user was represented as a userId. Ratings are on a scale of 1 to 10.}
  \label{tab:groupscenario}
  \begin{tabular}{cccccc}
    \toprule
    UserId & Item\_1& Item\_2&Item\_3&Item\_4&Item\_5\\
    \midrule
    user\_57749 & 4 & 2 & 2 & 10 & 9 \\
    user\_78033 & 10& 7 & 3 & 4 & 7  \\
    \bottomrule
  \end{tabular}
\end{table}

\subsection{Pipeline}\label{sec:pipeline}
The upcoming paragraphs describe the pipeline used in the study to apply social choice-based aggregation strategies using LLMs. We describe prompt construction, LLM implementations and evaluation procedure. The entire pipeline for a group scenario is visualized in Figure \ref{fig:pipeline}. We iterated through the pipeline for every scenario in the full group database.

\begin{table*}
  \caption{All LLMs included in this study.}
  \label{tab:LLM}
  \begin{tabular}{cccl}
    \toprule
    Short model name & Number of parameters& Size&Name of implementation (+ Quantization) \\
    \midrule
    Mistral & 7B & 4.1GB&\textit{mistral\:instruct} (Q4\_0)\\
    Llama & 8B & 8.5GB&\textit{llama3.1:8b-instruct-q8\_0} (Q8\_0)\\
    Gemma & 9B & 5.4GB&  \textit{gemma2} (Q4\_0)\\
    Phi & 14B &9.1GB& \textit{phi4} (Q4\_K\_M)\\
    \bottomrule
  \end{tabular}
\end{table*}

\subsubsection{Prompt Construction}
For each group scenario, the prompt template was supplemented with the specific scenario information. First, one of the four social choice-based aggregation strategies was randomly selected (Table \ref{tab:strats}). Due to the fact that for each strategy four LLMs would need to generate a response, the time investment of applying all strategies to all groups was too high. By randomly selecting a strategy per group, we maximize the number of unique groups included in the study. The strategy was represented in the form of a social choice-based explanation, as used and discussed in previous work \cite{tran2019towards,barile2023evaluating}. The group table was formatted as a JSON dictionary containing per-item lists of ratings. The baseline data format grouped ratings per item as opposed to per user.  Each model received the exact same information. The prompt included instructions to output a recommendation list of items in case of a tie and JSON parsing, resulting in a dictionary with \textit{strategy} and \textit{recommendation} keys.\footnote{Full prompts are found in the repository.}

\subsubsection{LLMs}
To maximize reproducibility and simplicity in applying LLMs for group recommendation, we make use of \textit{Ollama} and the \textit{langchain} python package.\footnote{Langchain version 0.3.2} The included models are relatively small (\textit{quantized implementations}), thus can run locally, omitting the need to work with APIs or external servers. These local models minimize the privacy concerns within the Recommender System community as well as computing costs for researchers and practitioners. We generated recommendations using different models to showcase potential divergence in performance. The included models were \textit{Mistal}, \textit{Llama3}, \textit{Gemma2} and \textit{Phi4}. The exact implementations alongside their quantization are summarized in Table \ref{tab:LLM}. For each scenario, all four models were called in succession and generated recommendatons. 

\subsection{RQ1: Baseline Performance}
For each group scenario and corresponding, randomly chosen social choice-based aggregation strategy, the correct response was generated using a direct implementation of the strategy (\textit{gold\_label})\footnote{All implementations of social choice-based strategies are found in the Github repository: \url{https://github.com/Cwaterschoot/LLMs_SocialChoiceAggregation}}. To evaluate correct application of a strategy by the LLM, we opted to investigate the ability to derive the correct single item (top recommendation). If a strategy was correctly applied, the LLM should return the identical top recommendation as derived by applying the strategy itself\footnote{The evaluation of ranking capabilities by LLMs is introduced in a later section.}.

Before calculating accuracy, we derived the overlap between the gold\_label and the output of each of the four LLMs. The gold\_label was a single item (or list with a tie result) derived by the same social choice-based strategy that the LLM was prompted to apply. Certain social choice-based strategies, especially Least Misery (LMS) and Most Pleasure (MPL), tend to result in long lists of tie results at high degrees of group complexity. Since in practice only one item is recommended to a group, we evaluated an LLM-generated recommendation as correct if there was overlap between the gold\_label and LLM output. For example, if the correct result in applying the MPL strategy is \textit{[`item\_1', `item\_3']} and the LLM returned the recommendation as \textit{[`item\_3']}, we counted it as a correct. While we acknowledge that this is a soft definition of accuracy, as it ignores ties, this does reflect a correct result. We applied this definition of a correct result for all aggregation strategies.

\subsection{RQ2: Additional Prompt Conditions}\label{sec:conditions-methods}
Alongside the standard pipeline (baseline) outlined above, we investigated whether three adjustments to the prompt impacted performance compared to the baseline: 
 (i) prompting for natural language explanations, (ii) implementing in-context learning and, (iii) adding (real-world) domain cues to the group scenarios. For each of these conditions, we generated group recommendations using \textit{Phi4} for all groups consisting of $50$ items ($n=250$). Applying these modifications using all LLMs was not feasible within the timeline of this research. Therefore, we opted for the largest LLM (14B parameters). These group scenarios represent the highest degree of group complexity at $100$ (2 members), $200$ (4 members) and, $400$ (8 members). Full prompts are available in the companion repository.\footnote{\url{https://github.com/Cwaterschoot/LLMs_SocialChoiceAggregation}} All three conditions started from the basic prompt used in Section \ref{sec:pipeline} and were not built on top of each other.

\subsubsection{Explanations}
Generating natural language explanations is a popular use case for LLMs \cite{Lubos2024llmgen,Said2025explaining}. Thus, we investigated whether this task impacts the ability of the model to apply the social choice-based strategy correctly. We adjusted the output parsing to include a third key in the JSON output for the explanation itself. We added instructions to the prompt by requiring an explanation to the group represented in the scenario:
\textit{``Provide a short explanation detailing how you derived the recommendation. Explain to the group how the strategy works and why the output is being recommended to them.''}

\subsubsection{In-Context Learning}
Originally, Few-shot Learning (FSL) was applied to supervised machine learning and was implemented by only presenting a limited number of training examples (representing all classes) with supervised labels to the model \cite{Song2023acompr,Wang2021general}. In the case of LLMs, few-shot learning has been categorized under the umbrella of ``\textit{In-Context Learning}'', the ability of LLMs to make predictions based on a context containing a few examples, performing specific tasks without updating any parameters \cite{dong2024survey}. 

Thus, in our case, In-Context Learning was implemented via Few-shot prompting, providing the LLM with several group scenarios and the correct recommendation given the queried social choice-based aggregation strategy (see e.g. \cite{Petruzzelli2024instruct} for a similar approach regarding LLM recommendations). We generated three additional group scenarios with $50$ items each, representing all potential group sizes ($2,4$ and $8$). In each iteration, we applied the randomly chosen strategy (and only that strategy) to all three groups and added both group ratings and correct output to the prompt: ``\textit{If the input would be [group table], the correct recommendation would be [correct output].}'' All three examples were shown regardless of the group complexity of the prompted scenario.

\subsubsection{Domain Cues}\label{sec:domaincuesmethod}

As discussed in Section \ref{sec:background}, group recommendations are applied in various domains. To test whether real-world domain cues impacted performance, we compared output generated using anonymous group scenarios (baseline) with group scenarios in which the item identifiers (\textit{item\_x}) were replaced with movie titles. We randomly sampled $50$ movie titles from the Movielens dataset \cite{Harper2016movie} and used these titles as item names. Additionally, we changed every mention of \textit{item} in the prompt to \textit{movie}. No further adjustments to the group scenarios were made. 

\begin{figure*}[h]
  \centering
  \includegraphics[width=1\linewidth]{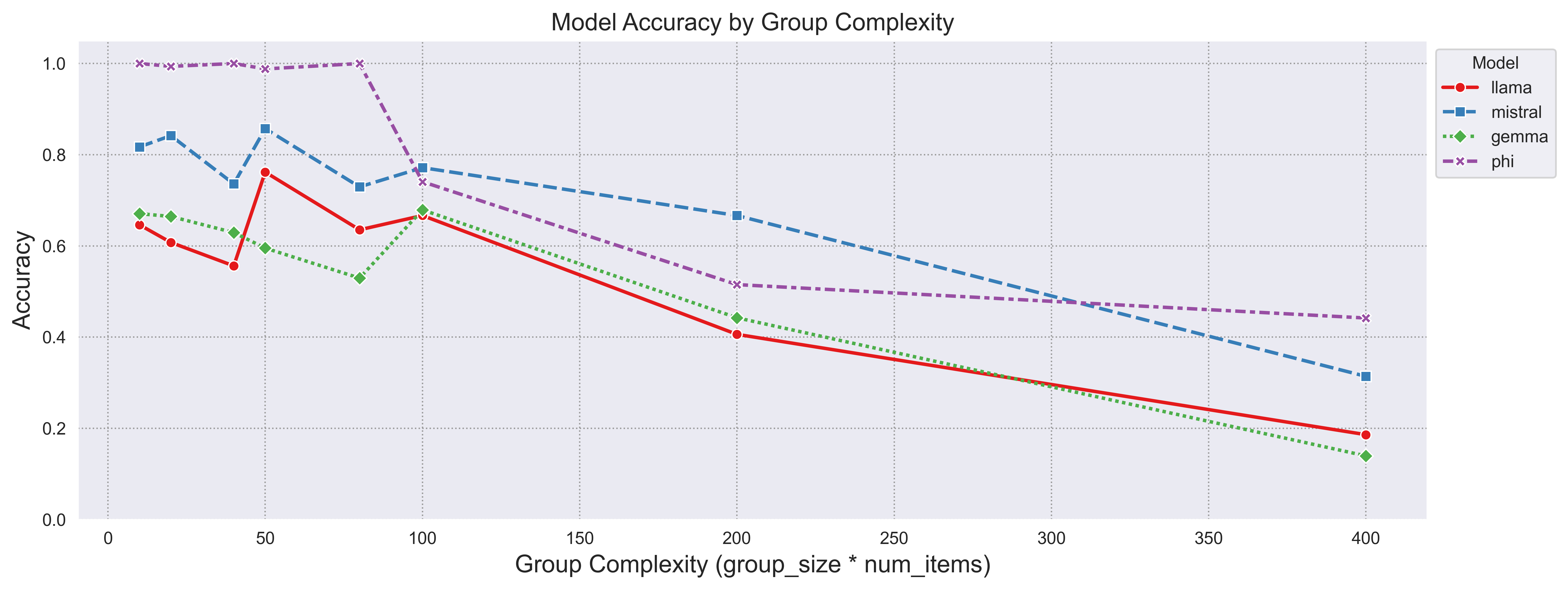}
  \caption{Performance (Accuracy) of all LLMs (Llama, Mistral, Gemma and Phi4) based on 1,000 group scenarios with varying degrees of group complexity. Complexity ranges from 10 (2 members x 5 items) to 400 (8 members x 50 items).}
  \label{fig:full-performance}
  \Description{Performance (Accuracy) of all LLMs (Llama, Mistral, Gemma and Phi4) based on 1,000 group scenarios with varying degrees of group complexity (number of ratings). Complexity ranges from 10 (2 members x 5 items) to 400 (8 members x 50 items). Accuracy decreases across the board when the number of ratings increased}
\end{figure*}

\subsection{RQ3: Impact of data formatting}
LLMs generate output based on a provided prompt and, as a result, are sensitive to the format in which the group scenarios are provided, as it becomes part of the prompt. During our initial testing, we found varying performance based on the format in which the group ratings were added to the baseline prompt. To further illustrate how the data formatting affects performance, we ran the subset of group scenarios presented in Section \ref{sec:conditions-methods} using the baseline prompt and \textit{Phi4}. We modified the data format in which the group scenario was presented: a JSON object containing lists of ratings per item (JSON\_item, used previously), a JSON object in which each user is represented as a dictionary containing their ratings (JSON\_user) and a standard dataframe created with the \textit{Pandas} python package, with users per row and items as columns.

\subsection{Explorative Analysis of Ranked Recommendations}
Previously, we discussed the evaluation of LLM-generated group recommendations using accuracy, disregarding the ranking of items. However, in a practical context, Recommender Systems often provide a ordered list of items and thus, performance metrics need to be adapted to investigate whether LLMs provide relevant items in their top 10. For this reason, we extended our evaluation and compared our previous performance with that achieved by evaluating ranked lists of items provided by LLMs.

We prompted \textit{Mistral} (the smallest model) and \textit{Phi4} (the largest model) to output the top 10 items calculated using the social choice-based aggregation strategy, as opposed to only the winning item(s) based on the strategy. Prompting the smallest and largest models in the study provides a clear picture of the ranking capabilities using the procedure described by an aggregation strategy. We used the baseline prompt without any additional conditions. We iterated through the 50 item group scenarios introduced in Section \ref{sec:conditions-methods} and calculated Normalized Discounted Cumulative Gain (NDCG) based on the top 5 (NDCG@5) and top 10 (NDCG@10) items recommended by the two LLMs. The NDCG takes the ranking into account, rewarding correctly recommended items higher up in the top ranked items in the list \cite{Wang2013atheoretical}.

\section{Results}\label{sec:results}
In the following section, we report the results obtained by applying social choice-based aggregation strategies using language models. First, we present the distribution of group scenarios based on complexity. Afterwards, we outline the performance by language models. Finally, we report the performance under several additional testing conditions, including prompting for an explanation of the procedure, in-context learning and adding domain cues to the data.

\subsection{Group distribution}
Following the procedure outlined in Section \ref{sec:groupgen}, we generated a total of $1,000$ groups. The distribution is summarized in Table \ref{tab:groupdistribution}. Due to the fact that group complexity is calculated by multiplying the group size with the number of items, not every degree of complexity is equally represented. For example, the complexity level of $100$ contains groups with \textit{2 members x 50 items}, as well as \textit{4 members x 25 items}, while complexity level $400$ only contains the groups with \textit{8 members x 50 items}\footnote{The distribution of aggregation strategies were similar for each group member-item set size pair}.

\begin{table}[H]
  \caption{Data distribution of generated group scenarios; by group size (number of group members) and number of potential items to recommend.}
  \label{tab:groupdistribution}
  \begin{tabular}{ccccc|c}
    \toprule
     & 5 items & 10 items &25 items & 50 items & Total \\
    \midrule
     2 members&82 & 75& 84&80& 321\\
     4 members &83 & 89& 82&79& 333 \\
     8 members & 89& 85& 86&86 & 346\\
     \midrule
     Total & 254& 249&252 &245& 1000\\
    \bottomrule
  \end{tabular}
\end{table}

\subsection{RQ1: Baseline Performance}\label{sec:baseline}
Overall, \textit{Phi4} outperformed all other models when considering the full dataset. The accuracy of both \textit{Llama} and \textit{Gemma} was $0.56$. \textit{Mistral} achieved an accuracy of $0.73$, while \textit{Phi4} resulted in an accuracy score of $0.83$.

Figure \ref{fig:full-performance} provides a summary of the results based on group complexity. Overall, the performance of the included language models followed a similar trend as group complexity increased, albeit with varying accuracy scores. \textit{Gemma}, \textit{Llama} and \textit{Mistral} underperfomed at lower group complexity compared to \textit{Phi}, which maintained near-perfect accuracy up to a group complexity of 80 (groups with 8 members and 10 items). From a matrix size of 100 onward, all models showed a decline in performance (Figure \ref{fig:full-performance}). While \textit{Llama} and \textit{Gemma} experienced a steady decrease, \textit{Mistral} achieved the highest accuracy score for group complexity 100 and 200. Ultimately, \textit{Phi4} achieved the best performance at the highest complexity.

Furthermore, we found variation in performance across the four social choice-based aggregation strategies. When prompted to apply the Most Pleasure (MPL) strategy, the LLMs achieved the highest accuracy score of $0.82$. Approval Voting (APP) followed with a score of $0.69$, while Additive Utilitarian (ADD) reached an accuracy of $0.62$. Least Misery (LMS) proved to be the most challenging to apply correctly, with the LLMs reaching an accuracy of only $0.58$. \textit{Phi4} consistently outperformed all other models across all social choice-based aggregation strategy, followed by \textit{Mistral}. We found \textit{Llama} to be the worst when applying APP and LMS, while \textit{Gemma} had the lowest accuracy score for ADD and MPL.

\subsection{RQ2: Additional Prompt Conditions}
\begin{figure}[h]
  \centering
  \includegraphics[width=1\linewidth]{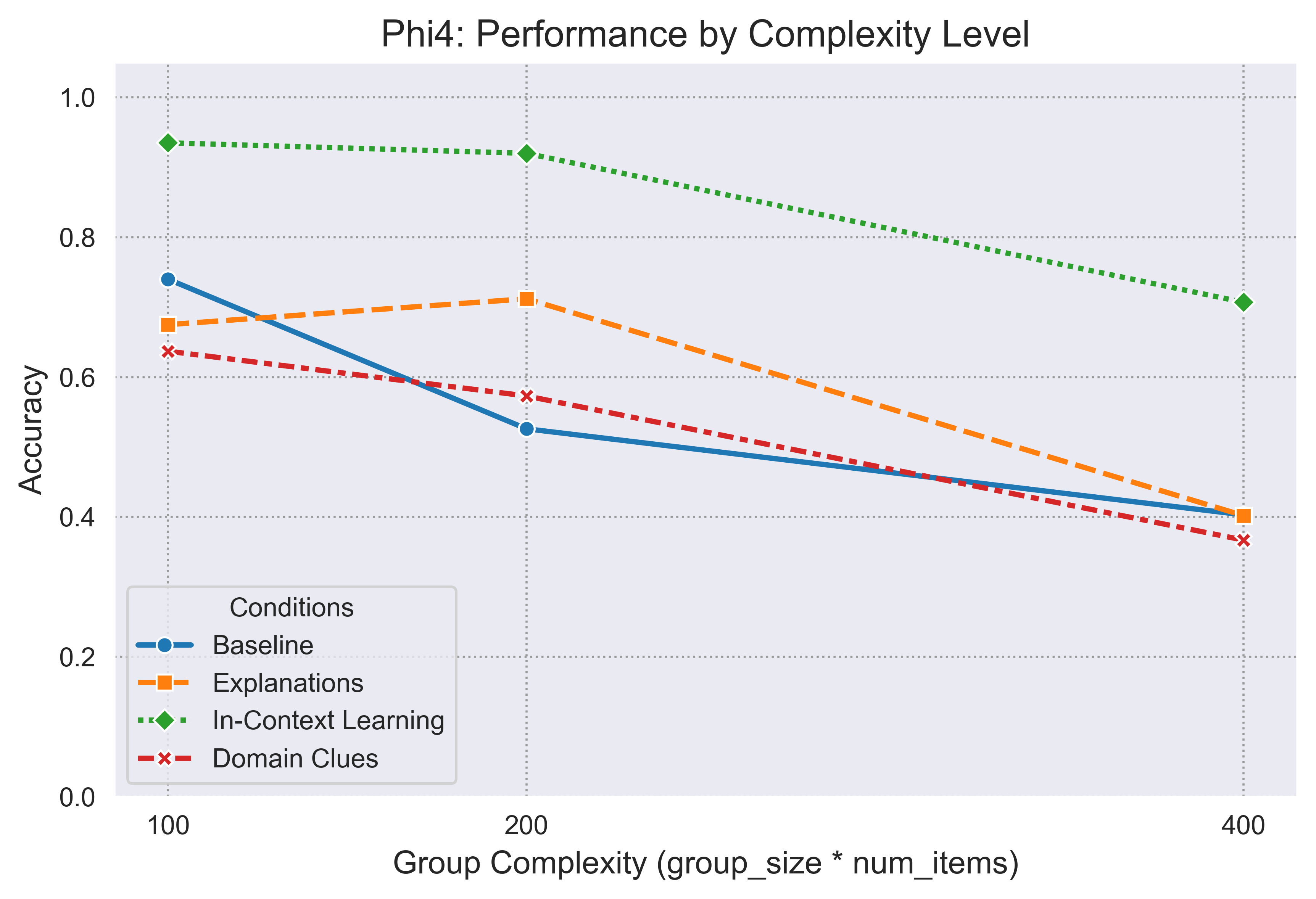}
  \caption{Performance of Phi4 based on prompt adjustments: (i) requiring explanations, (ii) implementing in-context learning and, (iii) adding domain cues. Group complexity was set at 100, 200 and 400 ratings (50 items; 2, 4 or 8 group members).}
  \label{fig:conditions}
  \Description{Performance by Phi4 under the different conditions compared to the baseline. We prompted for natural language explanations, in-context learning and we added domain cues in the form of movie titles.}
\end{figure}

As described in Section \ref{sec:baseline}, although \textit{Phi4} was the best performing model, its accuracy steadily declined at complexity levels above 80. The following paragraphs detail the performance of \textit{Phi4} at group complexities of 100, 200 and 400 (scenarios with 50 items, $n=245$). The aim was to investigate whether prompt modifications or adjustments to the group scenario improved model performance in correctly applying social choice-based aggregation strategies when provided with large rating matrices. Figure \ref{fig:conditions} summarizes the accuracy of \textit{Phi4} under the studied conditions at higher group complexity. For group complexity 100, 200 and 400 and prompted with the baseline prompt used in Section \ref{sec:baseline}, \textit{Phi4} achieved an accuracy score of $0.57$.

\subsubsection{Natural Language Explanations}
The baseline prompt provided to \textit{Phi4} was supplemented with a requirement to provide a natural language explanation. This explanation was intended to clarify for the group how the applied strategy functioned for their specific scenario. Overall performance improved to $0.59$, representing a 2 percentage point increase. This increase was mainly derived from better performance at a group complexity of $200$ (Figure \ref{fig:conditions}). However, at the highest level of group complexity, we observed no difference between the baseline and the prompt that additionally requested an explanation to the group.

\subsubsection{In-Context Learning}
The condition of In-Context Learning added three examples to the prompt, covering the spectrum of complexity levels that the LLM had to process. This approach typically improves model performance, and in this case, the inclusion of three examples significantly improved the model's ability to accurately apply social choice-based aggregation strategies. As a result, the overall accuracy score (at higher complexity levels) of \textit{Phi4} rose to $0.85$, an increase of 28 percentage points. In-Context Learning expanded the range of group complexity at which \textit{Phi4} maintained an accuracy score above $0.90$ (achieving $0.93$ at a group complexity of 100 and $0.92$ at 200). Accuracy dropped to $0.71$ only at a group complexity of 400 ratings (Figure \ref{fig:conditions}).

\subsubsection{Domain Cues}
The final condition adjusted the item names to actual movie titles derived from the Movielens dataset. We investigated whether adding a real-world domain to the data and prompt influenced the performance. Overall model performance at higher group complexities decreased with 4 percentage points, yielding an accuracy of $0.53$. Only at a complexity of 200 did we find a slightly better performance by adding domain cues to the prompt and group scenario (Figure \ref{fig:conditions}).

\subsection{RQ3: Impact of data formatting}\label{sec:dataformat}
To investigate whether the accuracy of LLMs was impacted by the format of the group scenario and user ratings, we analyzed three different formats (Table \ref{tab:dataformatting}).
The data format which lists ratings per item achieved the highest accuracy. Importantly, formatting the group scenario on a user basis, i.e. rating lists per user, achieved a lower accuracy. The differences presented in Table \ref{tab:dataformatting} show the importance of data formatting when using LLMs to generate group recommendations.

\begin{table}[H]
  \caption{Accuracy scores of \textit{Phi4} using the baseline prompt at increasing group complexity (100, 200 or 400 ratings); group ratings formatted either as JSON object containing per-item lists (JSON\_item), JSON object containing per-user entries (JSON\_user) or a Pandas dataframe (Dataframe). }
  \label{tab:dataformatting}
  \begin{tabular}{cccc}
    \toprule
    & JSON\_item & JSON\_user & Dataframe\\
    \midrule
    100 ratings &0.74 &0.59 &  0.58\\
    200 ratings & 0.53 & 0.24& 0.45\\
    400 ratings & 0.40 & 0.16& 0.40\\
    \bottomrule
  \end{tabular}
\end{table}

For GRS applications using LLMs, we argue it is crucial to include data formatting details in the methodology to guarantee reproducibility. Furthermore, future work that incorporates group scenario data should perform initial testing using various formats to investigate performance differences and ensure this factor is optimized during the experimentation and evaluation phases.

\subsection{Explorative Analysis of Ranked Recommendations}

\begin{table}[H]
  \caption{Average NDCG scores of \textit{Mistral} and \textit{Phi4} for group scenarios with 50 items ($n=245$); by increasing group complexity (100, 200, or 400 ratings)}
  \label{tab:ndcg}
  \centering
  \begin{tabular}{ccccc}
    \toprule
    & \multicolumn{2}{c}{Mistral} & \multicolumn{2}{c}{Phi4} \\
    \cmidrule(lr){2-3} \cmidrule(lr){4-5}
    & nDCG@5 & nDCG@10 & nDCG@5 & nDCG@10 \\
    \midrule
    100 ratings & 0.65&0.78 & 0.93& 0.97\\
    200 ratings & 0.54& 0.72& 0.76& 0.83\\
    400 ratings & 0.42& 0.67& 0.57& 0.74\\
    \bottomrule
  \end{tabular}
\end{table}

Finally, we analyzed the top ranked items to investigate the ranking performance of two LLMs using the baseline prompt without any prompt modifications. More specifically, we prompted the smallest LLM (\textit{Mistral}) (7B parameters) and the largest included model (\textit{Phi4}) (14B parameters) to provide the top 10 and calculated NDCG@5 and NDCG@10, averaged across 245 groups (Table \ref{tab:ndcg}). 

Following earlier trends, performance decreased when the number of ratings increased. Unsurprisingly, \textit{Phi4} outperformed \textit{Mistral} once more, achieving an NDCG@10 of $0.97$ when processing 100 ratings (Table \ref{tab:ndcg}). However, the ranking capabilities starkly decreased at 400 ratings, for which \textit{Phi4} achieved an average NDCG@10 of $0.74$. This results illustrated once more the extent that group complexity affects LLM performance in correctly applying social choice-based aggregation strategies. \textit{Mistral} achieved an optimum NDCG@10 of $0.78$ when processing 100 ratings, decreasing to an average NDCG@10 for 400 ratings of $0.67$.

\section{Discussion} 
In this study, we investigated whether LLMs can be prompted to correctly apply social choice-based aggregation strategies. In the following section, we contextualize our results. We argue in favor of including a varying degree of group complexity as well as different data formats in the experimental pipeline of GRS. Additionally, we discuss the limitations of the current study and how future work can address these shortcomings.

\subsection{The Impact of Group Complexity}
Our results indicate that group complexity, defined as the product of group size and number of items, is an important factor in the evaluation of GRS capabilities of LLMs. \textit{Phi4} achieved near-perfect accuracy scores up until a group complexity of 80 (Figure \ref{fig:full-performance}). Other included LLMs were also found to be quite stable up until that point. However, as group complexity increased, performance of LLMs decreased as well. While not all included LLMs were equally sensitive to this factor, a general decrease in performance was found starting at 100 ratings. Prompting for a natural language explanation or adding domain cues to the group scenarios and prompt did not affect performance in a significant way. 

As a result, evaluating LLM-generated recommendations derived by providing a small set of ratings may lead to an incomplete picture of the performance of the model. Therefore, we argue in favor of including varying degrees of group complexity in the evaluation of LLM-generated recommendations by generating groups with a varying number of group members and increasing list of items.

\subsection{Using Smaller Models}

Privacy concerns accompany the use of LLMs \cite{Kibriya2024privacy}. A key issue is the inability to ``unlearn'' personal data, which is especially relevant when developing applications that generate output based on user information, such as recommender systems \cite{Kibriya2024privacy,Jeckmans2013privacy}. One way to address part of these concerns is the use of smaller, open-source models which can be run or fine-tuned locally \cite{Wiest2024privacy,Hou2024finetuning}. Additionally, smaller LLMs eliminate the reliance on APIs, which comes with risks such as model availability, price fluctuations or even discontinuation. They also require fewer computing resources, presenting an efficient alternative to larger models. If local LLMs yield strong results on a particular task such as applying social choice-based aggregation strategies, there is a compelling case to opt for them instead of larger models.

\subsection{Over-recommending}\label{sec:over-recommending}
Due to the use of overlap to calculate accuracy, a model benefits from over-recommending, i.e. returning a longer list of recommendations than needed. We find that only \textit{Mistral} over-recommended items (Table \ref{tab:listlength}). The ground\_truth recommendation list length derived by the social choice-based aggregation strategies themselves was $2.87$, while \textit{Mistral} returned on average $3.80$ items. This result might partially explain the performance of \textit{Mistral} at higher complexities (Figure \ref{fig:full-performance}). On the other hand, \textit{Phi4} did not over-recommend and approximated the ground\_truth length with an average list of $2.36$ recommended items. Thus, for \textit{Phi4}, we conclude that using overlap as procedure for calculating accuracy was not problematic due to the lack of over-recommending.

\begin{table}[H]
  \caption{Mean length (and standard deviation) of recommendation list using the baseline prompt. Ground\_truth refers to the social choice-based aggregation strategies.}
  \label{tab:listlength}
  \begin{tabular}{ccccc}
    \toprule
    Ground\_truth & Llama &Mistral & Gemma & Phi4 \\
    \midrule
    2.87&1.56 & 3.80& 2.02&2.36\\
    (4.23) & (2.49) & (4.60) & (2.72) & (2.96) \\
    \bottomrule
  \end{tabular}
\end{table}

\subsection{Group configurations}\label{sec:groupconfigs}
The literature outlines that GRS need to be adapted to the group configuration itself, as it may affect the effectiveness of recommendations \cite{delic2020effects,barile2023evaluating}. Previous work by \citet{barile2023evaluating} has introduced several group configurations, different group compositions of users calculated on the basis of (dis)similarity between preferences: \textit{uniform} (high similarity among group members), \textit{divergent} (low similarity), \textit{coalitional} (two distinct sub-groups) and \textit{minority} (high similarity among group members with the exception of one single member). At high group complexity, these configurations might be derived using correlation metrics \cite{baltrunas2010group}. It is unclear whether these configurations will impact the performance of LLMs in applying social choice-based aggregation strategies. With a growing drive towards personalization, also for groups, we will include the factor of group configuration in future work, investigation whether the capability of LLMs are impacted by (dis)similarity among users.

\subsection{Limitations}
We identified several limitations to our approach which have to be taken into account. First, we only generated random groups, without making the distinction between similar or dissimilar users. Nevertheless, random groups are a good baseline and starting point. When generated in large numbers, it will include a variety of similar and dissimilar groups by default. However, future work ought to make the explicit comparison between types of groups. The framework of group configurations discussed in Section \ref{sec:groupconfigs} can address this limitation in future work.

Second, to test the impact of domain cues, we only made use of movie titles derived from the Movielens dataset. it remains to be seen whether other domains would have resulted in the same outcome. Popular domains in the literature include tourism \cite{chen2021attentive}, music \cite{najafian2018generating} and restaurants \cite{barile2021toward}. A cross-domain analysis investigating multiple domains can test whether the lack of positive impact on LLM performance for GRS holds across multiple domains.

Certain strategies will often result in ties, meaning that multiple objects can correctly be recommended. For a recommender system, providing one of the correct items to the group is seen as desired model behavior. Thus, we made use of overlap to calculate accuracy, meaning that the LLM was correct if it provided at least one of the accurate items. To make sure the model did not simply output many items, we checked over-recommending in Section \ref{sec:over-recommending}. This operationalization can be seen as the implementation suitable for practical contexts. Making use of a stricter definition of accuracy, requiring LLMs to output all items provided by the strategy, will significantly impact the results by lowering accuracy. 

\section{Conclusion}
In this study, we investigated the ability of smaller LLMs, which can run locally, to apply social choice-based aggregation strategies correctly. We specifically focused on group complexity by generating groups with a varying number of group members and items, resulting in the number of total ratings ranging between 10 and 400. Additionally, we introduced several conditions to test whether the performance of LLMs could be improved. 

Our results showed unequal performance across different models. Additionally, we found that the number of ratings had a negative impact on accuracy of the included LLMs. Starting from 100 ratings, performance decreased across the board. However, our results indicated that In-Context Learning can be used to significantly improve LLM performance for group recommendation based on social choice-based aggregation strategies. Other additional testing conditions, prompting for natural language explanations or adding domain cues to the input, did not improve performance.

We discussed several implications based on our results. We argued that future work should evaluate their applications based on varying degrees of group complexity to investigate whether GRS are robust against increasing the total number of ratings to account for. Additionally, we illustrated the impact of data formatting in LLM prompts. We encourage future work to perform initial testing with varying formats for the group scenario and ratings to tune the prompt and optimize performance.


\bibliographystyle{ACM-Reference-Format}
\bibliography{reprobib}

\appendix

\end{document}